\documentclass[oneside,onecolumn,pdflatex,sn-mathphys-num]{sn-jnl}

\usepackage{graphicx}%
\usepackage{multirow}%
\usepackage{amsmath,amssymb,amsfonts}%
\usepackage{amsthm}%
\usepackage{mathrsfs}%
\usepackage[title]{appendix}%
\usepackage{textcomp}%
\usepackage{manyfoot}%
\usepackage{booktabs}%
\usepackage{algorithm}%
\usepackage{algorithmicx}%
\usepackage{algpseudocode}%
\usepackage{listings}%
\usepackage{makecell}

\theoremstyle{thmstyleone}%
\theoremstyle{thmstyletwo}%

\theoremstyle{thmstylethree}%

\begin{document}

\title[Article Title]{TiROD: Tiny Robotics Dataset and Benchmark for Continual Object Detection}


\author*[1]{Francesco Pasti}\email{francesco.pasti@dei.unipd.it}
\author[1]{Riccardo De Monte}
\author[1]{Davide Dalle Pezze}
\author[1]{Gian Antonio Susto}
\author[1]{Nicola Bellotto}

\affil[1]{\orgdiv{Department of Information Engineering}, \orgname{University of Padua},\\
\orgaddress{
\city{Padua}, \postcode{35131}, \country{Italy}}}

\abstract{Detecting objects with visual sensors is crucial for numerous mobile robotics applications, from autonomous navigation to inspection.
However, robots often need to operate under significant domains shifts from those they were trained in, requiring them to adjust to these changes.
Tiny mobile robots, subject to size, power, and computational constraints, face even greater challenges when running and adapting detection models on low-resolution and noisy images.
Such adaptability, though, is crucial for real-world deployment, where robots must operate effectively in dynamic and unpredictable settings.
In this work, we introduce a new vision benchmark to evaluate continual learning strategies tailored to the unique characteristics of tiny robotic platforms.
Our contributions include: (i) Tiny Robotics Object Detection~(TiROD), a challenging video dataset collected using the onboard camera of a small mobile robot, designed to test object detectors across various domains and classes; (ii) a comprehensive benchmark of several continual learning strategies on different scenarios using NanoDet, a lightweight, real-time object detector for resource-constrained devices.
While data collection is performed on-device, training and evaluation are conducted offline to establish a reproducible benchmark.
Our results highlight some key challenges in developing robust and efficient continual learning strategies for object detectors in tiny robotics. Dataset and code are publicly available \url{https://pastifra.github.io/TiROD}}.

\keywords{Continual Learning, Object Detection, Tiny Robotics}

\maketitle

\section{Introduction}\label{sec:introduction}
Tiny Robotics~\cite{tinyNeuman} is an emerging research area with the potential to impact significantly various fields, including autonomous inspection and search \& rescue applications~\cite{insp,search}.
For tiny mobile robots, it is essential to visually detect objects in the surroundings to effectively operate in unknown and dynamic environments. However, these systems often face constraints in terms of sensor quality, computational power, memory, and energy consumption, making the development of robust object detection particularly challenging.

Continual Learning~(CL)and lifelong algorithms~\cite{cont,Shaheen2022}, which allow models to adapt and learn from new data without forgetting prior knowledge, are essential for deploying object detection systems in real-world scenarios. Most neural networks and other machine learning methods, indeed, suffer from catastrophic forgetting~\cite{forgetting}, i.e. the model loses previously learned information when trained on new data. This issue is particularly relevant for tiny robotics applications, where the capacity for storing and processing data is limited.

Several datasets and benchmarks have been proposed to evaluate object detection~\cite{COCO, VOC} and CL for robotics~\cite{loris,Kasaei2024}. However, these benchmarks do not fully capture the unique real-world challenges faced by tiny mobile robots, which operate under severe resource constraints and must adapt to dynamic environments.
Specifically, existing datasets often fail to account for low-cost sensors, limited computational power, and the need for continuous adaptation to new domains and classes in unstructured settings.

\begin{figure*}
    \centering
    \captionsetup{type=figure}
    \begin{subfigure}[b]{0.54\textwidth}
        \centering
        \includegraphics[width=\linewidth]{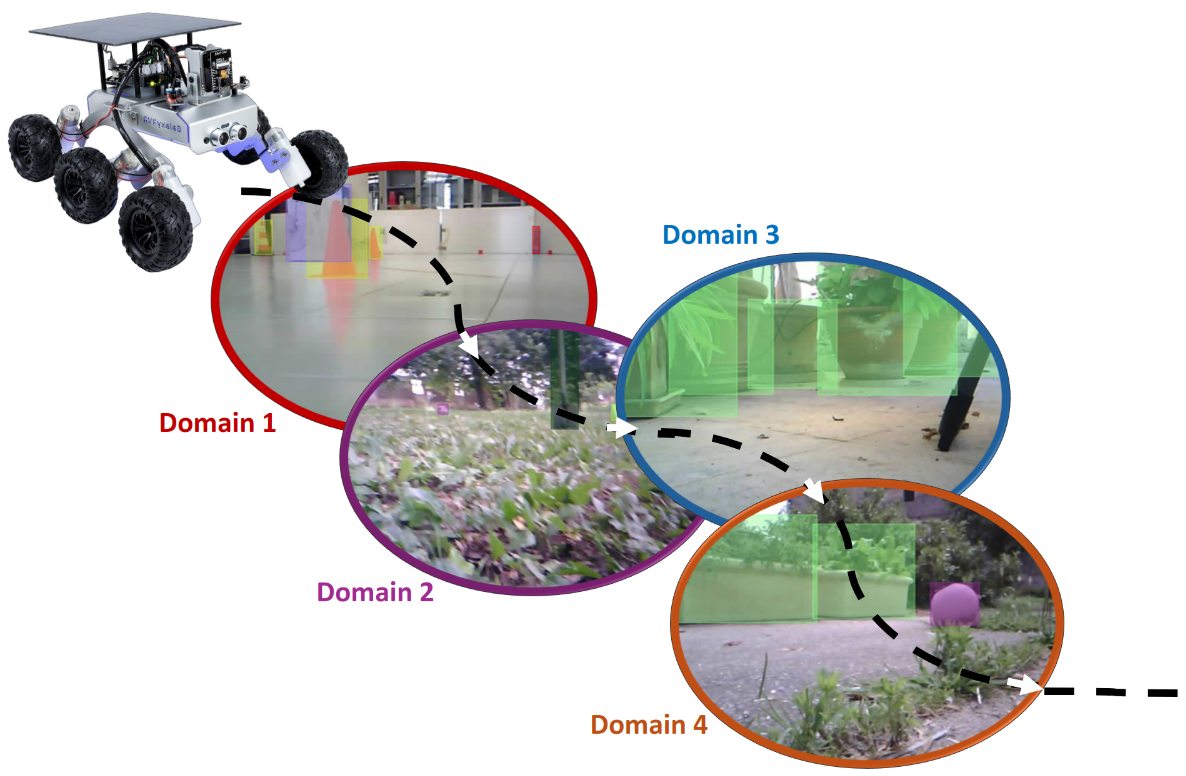} 
        \caption{\textit{TiROD} dataset and benchmark}
        \label{fig:tirod}
    \end{subfigure}
    \hfil
    \begin{subfigure}[b]{0.45\textwidth}
        \centering
        \includegraphics[width=\linewidth]{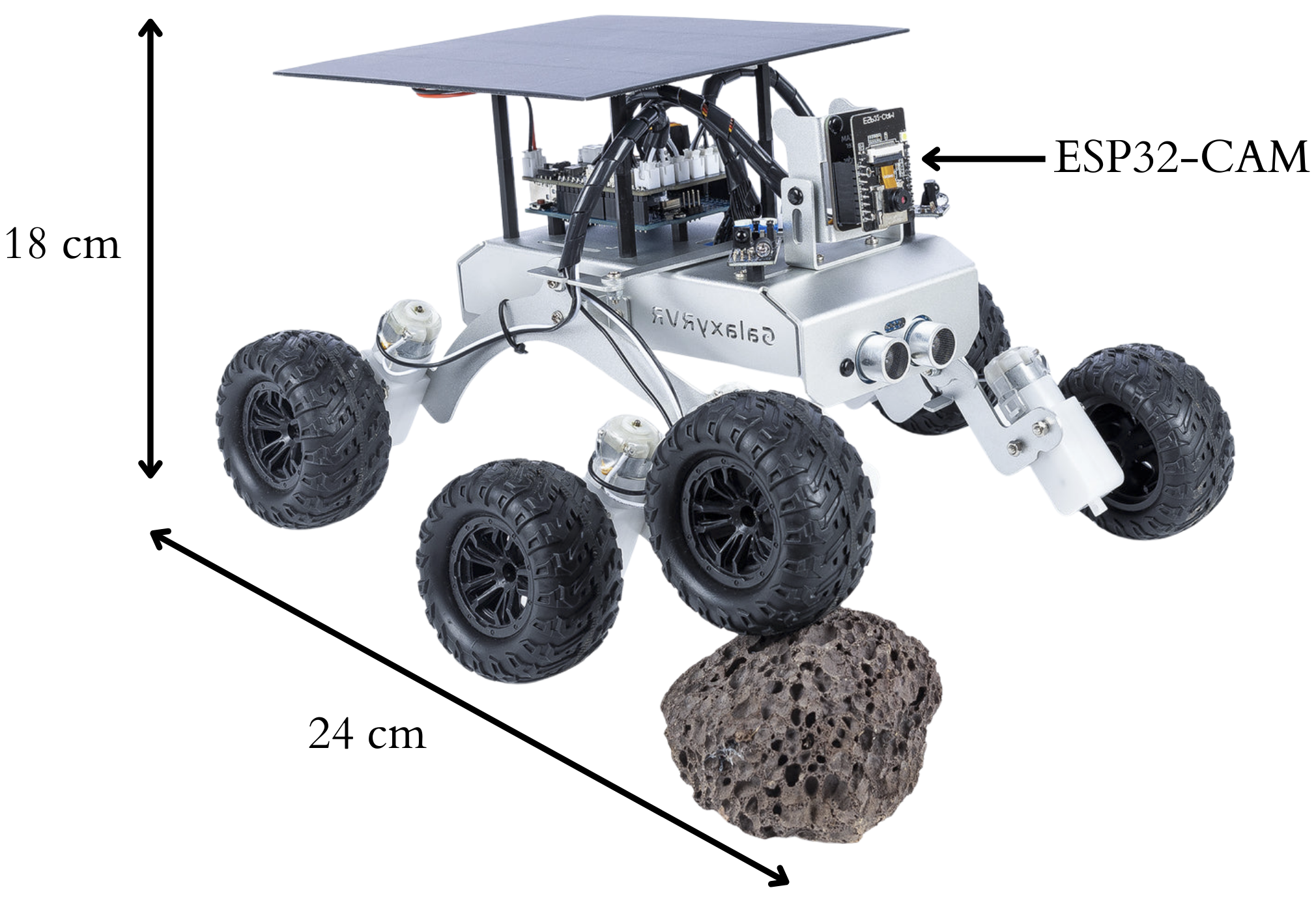}
        \caption{Tiny Robotic Platform}
        \label{fig:galaxyrvr}
    \end{subfigure}
    \caption{(a) \emph{TiROD}: Tiny Robotics Object Detection dataset, with a tiny mobile robot exploring multiple environments. The robot must learn to detect objects in each new domain without forgetting the previously acquired knowledge. (b) The Tiny Robotic platform used to acquire TiROD data via its onboard ESP32-CAM.}
    \label{fig:teaser}
\end{figure*}

To address these gaps, we propose a new vision dataset, Tiny Robotics Object Detection~(TiROD), and a benchmark designed to evaluate the performance of continual object detection for tiny robotics as illustrated in Fig.~\ref{fig:tirod}.
TiROD was collected using the low-cost onboard camera of a small mobile robot (Fig.~\ref{fig:galaxyrvr}), capturing 5 diverse unstructured environments (both indoor and outdoor) and 13 object classes. This dataset provides a challenging test bed for evaluating the adaptability of object detection systems under the constraints of tiny robotics.
While on-device training is an active area of research~\cite{lin2022device}, establishing the foundations for efficient Continual Object Detection (CLOD) on this type of data is an important prerequisite.
To this end, all CL training and evaluation in this work are performed offline, while the robot's onboard sensor is used exclusively for data acquisition.


Our work aims to facilitate the development of robust and efficient object detection systems for resource-constrained autonomous robots, enabling their deployment in challenging real-world environments.
To this end, our main contributions are:
\begin{itemize}
    \item TiROD, a new dataset collected using a tiny mobile robot, designed to test the performance of continual object detectors across different domains and classes;
    \item a benchmark of various CL strategies on two different scenarios with NanoDet~\cite{nanodet}, a popular lightweight object detector, providing valuable insights into their performance and limitations.
\end{itemize}

The data and the code are made publicly available\footnote[1]{\url{https://pastifra.github.io/TiROD}} to replicate the results and to foster further research in this field. We welcome and facilitate contributions on the project website to continuously expand the dataset across diverse domains, lighting conditions, and robotic platforms.

The remainder of the paper is structured as follows: Sec.~\ref{sec:tinyrobotics} summarises the state-of-the-art in Tiny Robotics, relevant Continual Learning for Object Detection~(CLOD) methods, and existing datasets; Sec.~\ref{sec:tirod_dataset} explains how we built the TiROD dataset and designed the CL scenarios; Sec.~\ref{sec:methods} presents the tested CLOD methods and the evaluation metrics, while Sec.~\ref{sec:experiments} illustrates and discusses the benchmark results including a computational overhead analysis in Sec.~\ref{sec:overhead}. Finally, Sec.~\ref{sec:conclusions} concludes the paper.

\section{Related Work}\label{sec:tinyrobotics}

There is a growing interest on research topics at the intersection between Tiny Robotics, Object Detection and Continual Learning. 
In this section we provide an overview of relevant works and discuss, in particular, the challenges of Continual Learning and Object Detection at the edge, including an analysis of existing datasets and benchmarks in these areas.

\subsection{Tiny Robotics}
Tiny robots are resource-constrained, low-cost, and lightweight autonomous systems~\cite{tinyNeuman}. Their potential capability to operate autonomously could significantly impact several fields, from agriculture to search \& rescue~\cite{TinyAgri,mcguire2019minimal}.
While there is no strict definition of {\em tiny robotics}, this term typically refers to small platforms whose primary computation relies on microcontrollers~(MCUs) or mobile CPUs rather than GPUs~\cite{tinyNeuman}. For these robots to operate autonomously, it is also vital that their power consumption is kept to a minimum.
The choice of algorithms and software design, therefore, must carefully addresses the resource limitations of tiny robots, imposing more stringent constraints than in traditional mobile platforms.

Recent progress in tiny robotics has focused on maximizing performance under tight hardware limitations., e.g. for Model Predictive Control~(MPC)~\cite{MPC} or Simultaneous Localization and Mapping~(SLAM)~\cite{SLAM}. To this end, Tiny Machine Learning~(TinyML)~\cite{TinyML} can contribute to the autonomy of such robots by enabling on-board data processing, reducing the need for continuous communication with external servers, and therefore saving bandwidth and power usage.

\subsection{Object Detection in Tiny Robotics}
Perception in small resources-constrained robots is notoriously challenging~\cite{Singh2024}. This paper focuses specifically on vision-based object detection~\cite{Obj}, which enables autonomous robots to perceive and interact with their environment. 
There has been a growing interest in deploying object detection at the edge, leveraging techniques such as depthwise-separable convolutions, neural architecture search, quantization, pruning, and knowledge distillation~\cite{zou2023object} to speed-up inference on constrained hardware~\cite{sah2024mcubench}.
Although these solutions enable object detection on resource-limited devices, the actual performance on a tiny robot is also limited by the motion-induced noise and the quality of its sensors. Indeed, low-cost cameras, commonly used in tiny robotics, often suffer from poor signal‑to‑noise ratios, motion blur, and other image degradation factors, leading to unique vision challenges~\cite{chen2008determining,steffens2021robustness}.
Our work addresses the latter by collecting real-world data, across multiple domains, with a tiny robot's camera.

\subsection{Continual Learning}
In the context of this paper, Continual Learning refers to the incremental update of a neural network model with new data without catastrophically forgetting previously acquired knowledge. According to~\cite{vanthree}, the literature considers three main scenarios: Task-Incremental Learning~(TIL), Domain-Incremental Learning~(DIL), and Class-Incremental Learning~(CIL).
In TIL~\cite{til}, models are trained on a sequence of tasks and evaluated on all of them. During testing, the model is provided with both the input sample and the task identifier. DIL~\cite{dil} instead is characterized by domain shifts in the task's data, while the set of labels to be detected remains the same across tasks. During testing, only the input sample is given, without additional information such as the task identifier. The same happens in CIL~\cite{cil}, where the model cannot access the task identifier during testing. Unlike DIL, though, each task in CIL introduces new classes, requiring the model to further expand its knowledge.
In this paper we target both DIL and CIL scenarios, as our tiny robot explores different domains that can also contain new classes.

CL methods can be categorized as (i)~rehearsal-based~\cite{replay,dgr}, (ii)~regularization-based~\cite{lwf, overcoming}, and (iii)~architecture-based approaches~\cite{schwarz2018progress, peng2023diode}. 
Rehearsal techniques store and reuse past data samples during training, with Experience Replay~\cite{replay1} being one of the most popular methods. Regularization approaches introduce additional constraints during training to preserve the memory of previous tasks, like penalizing model parameters based on their importance~\cite{overcoming} or using knowledge distillation to retain prior knowledge~\cite{lwf}. Finally, architecture-based approaches modify and expand the model's architecture to help maintain previously learned knowledge~\cite{schwarz2018progress}.

\subsection{Continual Learning for Object Detection}
Domain shift is a significant challenge for object detection in mobile robotics. Research in Domain Adaptive Object Detection aims to adapt models trained on labeled source domains to unlabeled target ones.
However, typical methods consider a single source–target pair, optimizing only for the target and ignoring catastrophic forgetting of the source~\cite{oza2023unsupervised}. Benchmarks are usually based on synthetic-to-real or weather variations in datasets such as Cityscapes, Foggy Cityscapes, and Sim10k~\cite{cordts2016cityscapes,sakaridis2018semantic,johnson2017driving}.
In contrast, mobile robots have to deal with perception sequences of dynamic environments and varying object categories, where past knowledge must be preserved. 
CLOD explicitly addresses this challenge~\cite{menezes2023continual}: models are updated with new tasks (defined by domains or classes) mitigating catastrophic forgetting, thereby maintaining sufficiently good performance across all previously seen tasks.

Most works in the CLOD literature use replay and regularization techniques to overcome the catastrophic forgetting problem.
Many solutions use distillation techniques on Fast or Faster-RCNN relying on a Region Proposal Network~(RPN) to get class-agnostic bounding boxes~\cite{Shmelkov_2017_ICCV, liu2020incdet}.
Rehearsal-based CLOD approaches often combine other CL techniques with replay to improve the results~\cite{acharya2020rodeo}.
With the growing interest in anchor-free and real-time object detectors, like recent versions of YOLO and FCOS architectures~\cite{jiang2022review, tian2020fcos}, some researchers are using these models to propose new replay- and regularization-based CL techniques~\cite{peng2023diode,monte2024rcplod,peng2021sid}.
Recently, there has been also significant interest in efficient CLOD solutions, targeting lightweight architectures and embedded systems~\cite{pastiLD}.

\subsection{Related Datasets}
Several datasets exist in the CLOD literature, as shown in Table~\ref{tab:datasets}. The most popular benchmarks are based on category recognition datasets like Pascal VOC~\cite{VOC} and Microsoft COCO~\cite{COCO}. 
However, these do not adequately reflect the challenges of mobile robot applications, where sensor data is typically not independent nor identically distributed, like the images of the mentioned datasets.
\begin{table}
    \centering
    \caption{Related datasets for Continual Object Detection.\label{tab:datasets}}
    \scriptsize
    \setlength{\tabcolsep}{2.5pt}  
    \begin{tabular*}{\textwidth}{@{\extracolsep\fill}|l|lllll|}
        \hline
        \textbf{Dataset} & \textbf{Context} & \textbf{Acquisition} & \textbf{Domains} & \textbf{CL Type} & \textbf{Focus}\\
        \hline
        CORe50~\cite{core} & {\bf Indoor, Outdoor} & Handheld & 2 & CIL & Classification \\
        F-SIOL-310~\cite{310}& Indoor & {\bf Robot} & 1 & FSIL & Classification\\
        OpenLORIS~\cite{loris} & Indoor & {\bf Robot} & 3 & DIL & Classification \\
        VOC2007~\cite{VOC} & {\bf Indoor, Outdoor} & Internet & - & CIL & {\bf Detection} \\
        COCO~\cite{COCO} & {\bf Indoor, Outdoor} & Internet & - & CIL & {\bf Detection}  \\
        OAK~\cite{OAK} & Outdoor & Human & - & {\bf DIL+CIL} & {\bf Detection} \\
        ROD~\cite{ROD} & Indoor & Turntable & 1 & - & {\bf Detection} \\
        ARID~\cite{ARID} & Indoor & {\bf Robot} & 1 & - & {\bf Detection} \\
        Active Vision~\cite{ammirato2017dataset} & Indoor & {\bf Robot} & {\bf 9} & - & {\bf Detection} \\
        CLAD-D~\cite{CLAD} & Outdoor & Car & {\bf 4} & DIL & {\bf Detection} \\
        \hline
        TiROD (ours) & {\bf Indoor, Outdoor} & {\bf Tiny Robot} & {\bf 5} & {\bf DIL+CIL} & {\bf Detection} \\
        \hline
    \end{tabular*}
\end{table}

Robotics-focused datasets include the Autonomous Robot Indoor Dataset~(ARID)~\cite{ARID}, the Robotic Object Detection~(ROD) dataset~\cite{ROD}, and the Active Vision Dataset~\cite{ammirato2017dataset}.
ARID was collected indoors using a wheeled mobile robot, with challenges such as occlusions and illumination changes. However, this dataset was not designed for CL applications and does not distinguish between different domains.
Similar limitations affect the ROD dataset, which was collected by a fixed RGB-D camera in a controlled environment (objects spinning on a turntable) without accounting for the challenges faced by real-world mobile robots. The Active Vision Dataset is tailored for instance recognition and, although not designed for CL, provides a clear distinction between domains, but it is limited to semi-static snapshots of simple indoor environments, therefore omitting the complexities of continuous video recordings in outdoor scenarios.
All these datasets have been collected using fairly high-end cameras in indoor settings.
TiROD, instead includes multiple indoor and outdoor sequences where the low-cost vision sensor is susceptible to motion blur, occlusions, and difficult lighting conditions, which better reflect the challenges faced by tiny mobile robots in real-world unstructured environments.

A popular dataset to evaluate CL in robotics is OpenLORIS~\cite{loris}, which was created using an indoor mobile robot with variations of illumination, clutter, occlusion, and object size in a DIL setting. Although collected in three different environments (home, office, and mall), these were not distinctly separated in the dataset, since the authors' main purpose was to isolate and evaluate only the aforementioned variations.
Another relevant dataset is F-SIOL-310~\cite{310}, which targets a challenging few-shot incremental learning problem in robotics, where only a few labeled examples are available to learn new classes. 
Our dataset, instead, emphasizes the domain and class variations that occur when a tiny mobile robot changes environment, offering an alternative CLOD evaluation tool for mobile robotics.
Moreover, both OpenLORIS and F-SIOL-310 focus on object recognition (i.e. image classification) or single-instance object detection, rather than multiple-instance like in TiROD.

Similarly, the CORe50~\cite{core} dataset, collected using a handheld camera, does not address the challenges of multiple annotated objects per frame, which in robotics are often out-of-centre, occluded, or blended into the background.
The Continual Learning Autonomous Driving~(CLAD)~\cite{CLAD} dataset, instead, is tailored for autonomous driving. It contains images collected by a car in various locations characterised by varying lighting and weather conditions. These are rather different though from the scenes observed by a tiny mobile robot. Moreover, CLAD relies on a high-quality camera setup, as opposed to the low-cost sensors of our robot. 
Finally, a challenging dataset is Objects Around Krishna~(OAK)~\cite{OAK}, which was collected by a walking person with an action camera to evaluate CL performance over time, analyzing the system's output every 15 frames. However, like for previous datasets, OAK's images and camera's perspective are rather different from those of a tiny robot. Also, OAK does not clearly distinguish tasks, and it is therefore unsuitable for evaluating CL on multiple domains.

The above datasets and their main details are summarised in Table~\ref{tab:datasets}. In contrast to them, TiROD is specifically designed to evaluate continual object detection on tiny mobile robots with low-cost cameras, combining domain and class incremental learning in indoor and outdoor environments. This complements and extends existing benchmarks by taking into account more challenging sensing and task assumptions.

\section{TiROD Dataset}\label{sec:tirod_dataset}
TiROD is a new dataset emphasizing the challenges of object detection and continual learning in real-world environments for Tiny Robotics. Here we describe the robot platform, the dataset details, and the considered scenarios.

\subsection{Tiny Mobile Robot}
TiROD was collected using a SunFounder GalaxyRVR rover~\footnote{https://docs.sunfounder.com/projects/galaxy-rvr}, which is a tiny mobile robot weighing 950g and measuring 24$\times$18$\times$10cm.
It is a low-cost, open-source solution suitable for outdoor terrains. Moreover, its built-in solar panel allows for extended operation in sunlit environments, making it ideal for many long-term navigation tasks.
The robot, shown in Fig.~\ref{fig:galaxyrvr}, is equipped with six TT gear motors and a Rocker-Bogie suspension system~\cite{bickler1993new} designed for challenging terrains.
The motors are controlled via drivers connected to an Arduino Uno board.
Additionally, the rover has an ESP32-CAM\footnotemark[2] MCU with a built-in 2.0 megapixels camera.

\subsection{Dataset Details}\label{sec:tirod_dataset_details}
The TiROD dataset was collected while driving our tiny robot across five different environments, one indoor and four outdoor, with progressively rougher terrains.
The first is an indoor environment with tiled floor. The second is outdoor, also with tiles. The third environment involves outdoor navigation across both concrete and grass. The fourth features longer sequences traversing grass. The final environment includes grass, foliage and dirt terrains.

\begin{figure}
      \centering
      \includegraphics[width=\textwidth]{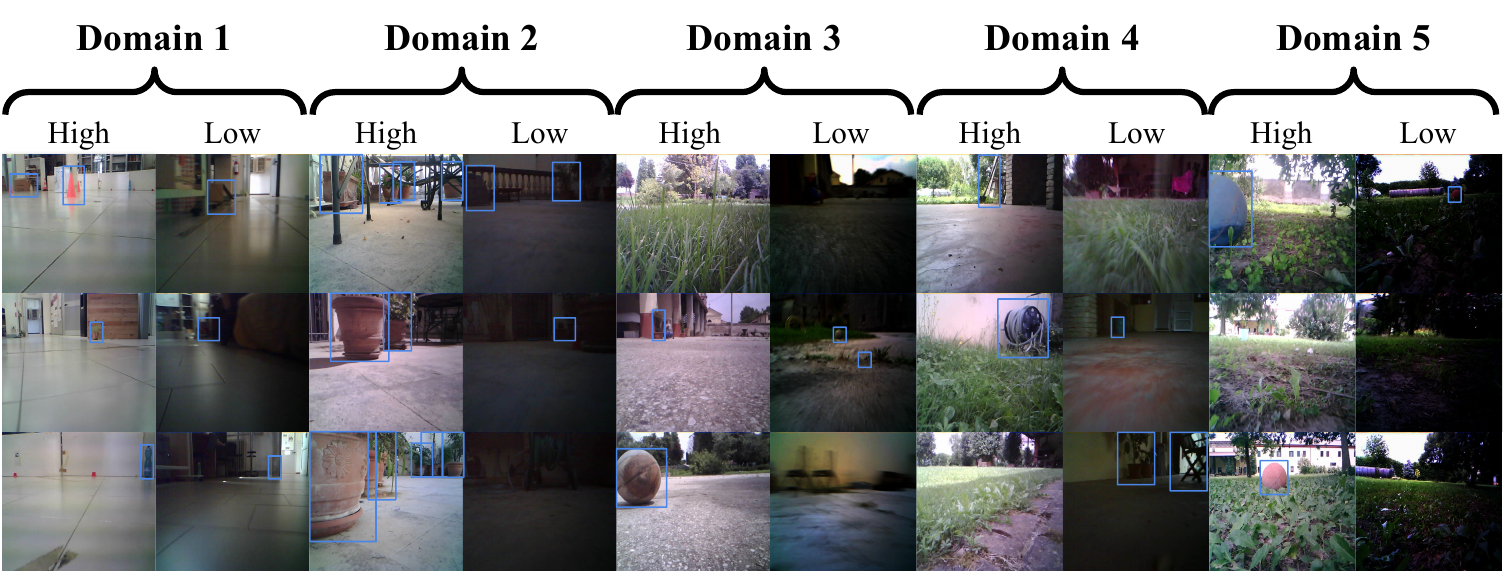}
      \caption{\emph{TiROD} dataset samples for the Cross Domain scenario.
      Each column corresponds to one of the 10 Continual Learning tasks.
      Every two tasks there is a domain change while for each domain there are two illumination conditions, ``High'' and ``Low''.\label{fig:dataset}}
\end{figure}

\begin{figure}
    \centering
    \includegraphics[width=1.01\textwidth]{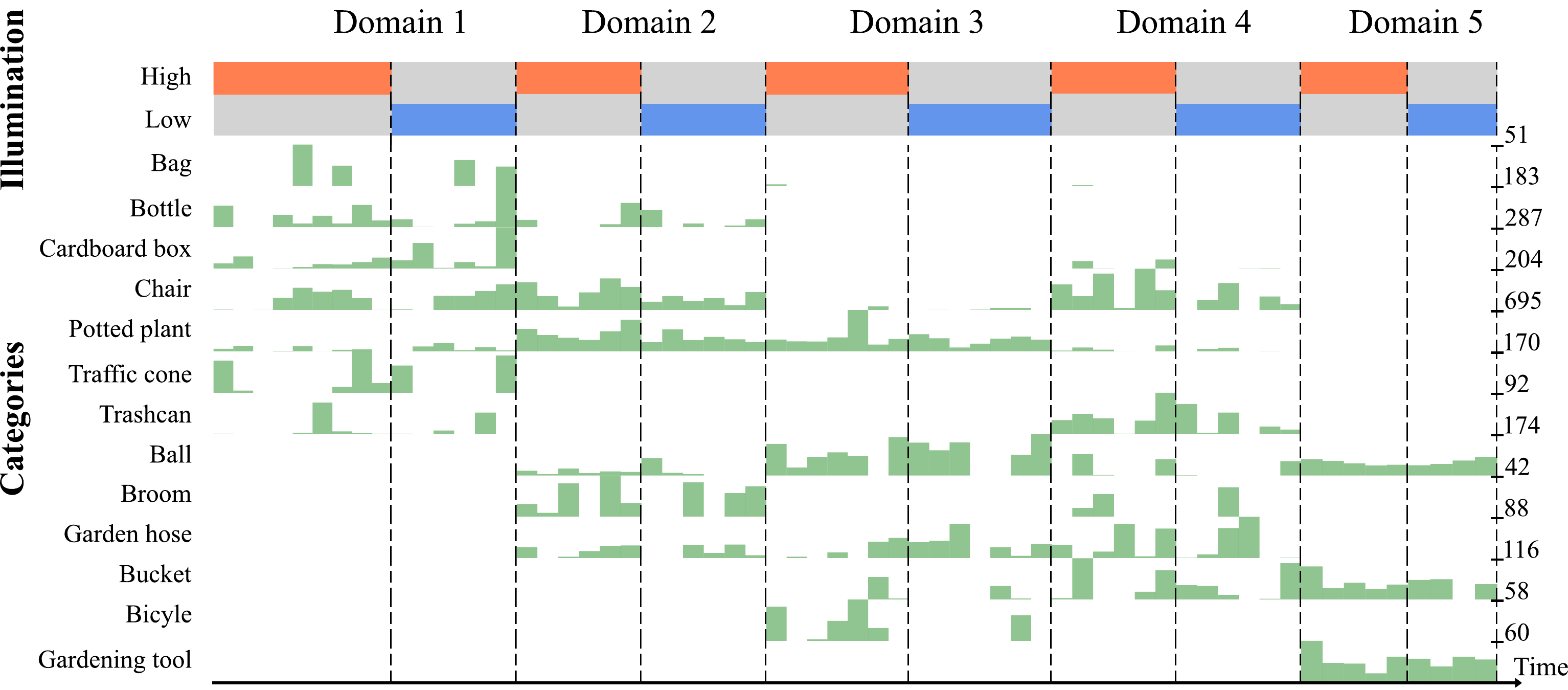}
    \caption{\emph{TiROD} full dataset statistics for the Cross Domain scenario. The x-axis represents time progression in 100-frame segments, with vertical dotted lines marking the separation between learning tasks. The robot was in continuous motion throughout most of each run; differences in the number of frames reflect the length of each trajectory. The top two rows represent the illumination level. Below, per-category histograms show the number of object instances observed in each 100-frame segment; the histograms' scales are shown on the right.\label{fig:datastats}}
\end{figure}

We captured RGB images at a resolution of 640$\times$480 pixels. The low-cost camera sensor and the robot's motion introduced significant noise and motion blur, due to fast direction changes and obstacles. These are additional challenges for real-world object detection with real mobile robots, particularly while they operate in unpredictable and unstructured environments.
We sampled images from the video stream at a frame rate of approximately 8~FPS.
For each domain, we collected data twice under two different illumination conditions, ``High'' and ``Low''.
This results in ten distinct tasks for our Cross Domain CLOD scenario, as depicted in Fig.~\ref{fig:dataset}.
The full dataset statistics, considering all the collected images, are instead reported in Fig.~\ref{fig:datastats}.

\begin{figure}
    \centering
    \includegraphics[width=\textwidth]{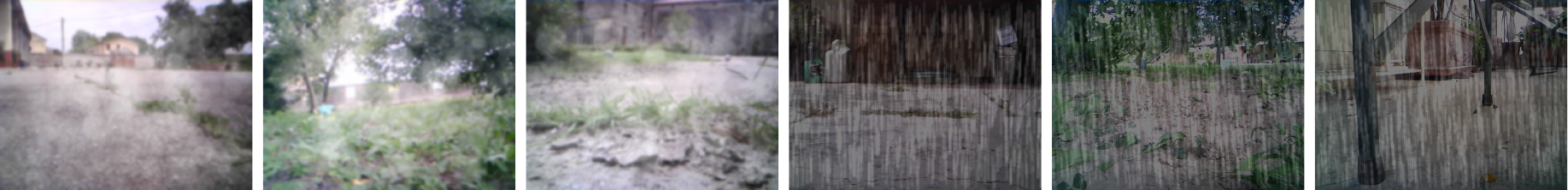}
    \caption{Examples of weather-based augmentations applied to the \emph{TiROD} dataset. The three images on the left demonstrate the fog augmentation, while the three on the right the rain augmentation. \label{fig:augmentations}}
\end{figure}

To increase dataset diversity and provide a controlled benchmark for studying CLOD under environment's appearance shifts, we applied weather‑based augmentations using the Albumentations library~\cite{albumentations} as depicted in Fig.\ref{fig:augmentations}. Specifically, we generated rain and fog variants of the outdoor images for Domains 2-5, expanding the dataset to a total of 17.4k images. 
Similarly to OpenLORIS~\cite{loris}, each sequence was randomly split with an 80/10/10 ratio into training, validation, and test sets. This splitting choice guarantees that all object classes are adequately represented in both the training and evaluation sets. Conversely, a strict temporal split would completely omit classes that are highly localized along the robot's trajectory (e.g., the `Bicycle' class appears only towards the end of Domain 3 High, as shown in Fig.~\ref{fig:datastats}). To objectively ensure that this choice did not introduce temporal leakage, we followed the quantitative methodology in~\cite{ramos2025data} and evaluated the cosine similarity between train, validation and test images using an ImageNet pre-trained ResNet model. The analysis confirmed that $\sim$90\% of the validation and test sets are visually distinct from the training data, with only a small leakage, primarily corresponding to segments where the robot was stationary or moving slowly over
visually uniform terrain. This shows that the continuous motion of the robot over unstructured terrains naturally mitigates static frame-to-frame redundancy, thereby preserving the integrity of the splits.

Initial annotations were generated using Grounding Dino~\cite{liu2023grounding} and then manually refined by the authors. This exhaustive quality-control step was critical to ensure strict labeling accuracy and bounding box consistency across all sequences, particularly given the unique visual challenges posed by the noisy, low-resolution images captured by the tiny robot's camera.

Table~\ref{tab:domain_counts} summarises the number of images per domain and illumination level, along with a short description of each domain. 
Table~\ref{tab:class_counts} reports the number of annotated object instances per category, as well as the number of images containing at least one instance. 
These numbers refer to the original images of TiROD, i.e., before applying the weather augmentations.
\begin{table}
\centering
\begin{minipage}{0.39\textwidth}
\caption{Details of TiROD dataset original images.}
\label{tab:domain_counts}
\footnotesize
\begin{tabular}{crrr}
\toprule
\textbf{Dom.} & \textbf{Illum.} & \textbf{Images} &\textbf{Description} \\
\midrule
1 & High & 914 &\multirow{5}{*}{} \\
 & Low & 680 & Indoor, Tiled Floor \\
\cmidrule(lr){1-4}
2 & High & 681 &\multirow{5}{*}{} \\
 & Low & 626 & Outdoor, Tiled Floor \\
\cmidrule(lr){1-4}
3 & High & 799 &\multirow{5}{*}{} \\
 & Low & 719 & Outdoor, Asphalt, Grass\\
\cmidrule(lr){1-4}
4 & High & 674 &\multirow{5}{*}{} \\
 & Low & 642 & Outdoor, Asphalt, Grass \\
\cmidrule(lr){1-4}
5 & High & 515 &\multirow{5}{*}{} \\
 & Low & 414 & Outdoor, Grass, Dirt\\
\bottomrule
\end{tabular}
\end{minipage}
\hfill
\begin{minipage}{0.39\textwidth}
\caption{TiROD per-category details.}
\label{tab:class_counts}
\footnotesize
\begin{tabular}{lrr}
\toprule
\textbf{Category} & \textbf{Objects} & \textbf{Images} \\
\midrule
Bag             & 135  & 132  \\
Ball            & 2132 & 1584 \\
Bicycle         & 190  & 164  \\
Bottle          & 922  & 720  \\
Broom           & 259  & 237  \\
Bucket          & 924  & 781  \\
Box             & 1038  & 704 \\
Chair           & 2696 & 1393 \\
Garden hose     & 807  & 720  \\
Garden tool     & 292  & 254  \\
Potted plant    & 7342 & 2565 \\
Traffic cone    & 640  & 417  \\
Trashcan        & 519  & 463  \\
\bottomrule
\end{tabular}
\end{minipage}
\end{table}

\subsection{Continual Learning Scenarios}\label{sec:tirod_dataset_benchmarks}

We consider two complementary scenarios to evaluate the performances of CLOD algorithms based on the data collected by our rover.

The first and more challenging one is the \textbf{Cross Domain} scenario defined by 10 sequential CL tasks and characterized by the 5 different environments explored by the robot under two illumination conditions. The tasks range from d1\_h~(read ``Domain 1, High illumination'') to d5\_l~(``Domain 5, Low illumination). The first two tasks correspond to an indoor environment, while the remaining ones are outdoor.
This scenario addresses the challenges of both DIL and CIL, requiring CLOD systems to adapt to significant domain shifts and varying data distributions.
The histograms in Fig.~\ref{fig:datastats} illustrate the distribution of object instances per category every 100 frames, highlighting the progressive emergence of new classes as the rover explores the environments.
Indeed, not all the categories are present from the first task; some only appear later~(e.g. ~\emph{Ball, Bicycle}) and may have few instances~(e.g.~\emph{Bag}).
Additionally, certain categories are outside the vocabulary of typical models~(e.g.~\emph{Garden hose, Gardening tool}), introducing more difficulties for pre-trained network adaptation.

To complement the previous one, we also define an \textbf{Intra-Domain} scenario using synthetic weather augmentations applied to Domain 5.
The task sequence includes clear weather conditions (high and low illumination), followed by foggy and rainy version of the same scene that were generated by the data augmentation described in  Sec.~\ref{sec:tirod_dataset_details}, resulting in 4 CL tasks. We selected Domain~5 for this scenario as its diverse ground texture, including a mix of dirt, foliage and grass terrains, makes it an interesting testbed.
This setup isolates the impact of environmental variation without introducing new classes, which allows us to evaluate how well the model retains knowledge and adapts to appearance shifts when deployed.

\section{TiROD Benchmark Setup and Methods}\label{sec:methods}
In this section we detail the experimental setup used to evaluate Continual Learning for Object Detection on the TiROD dataset. Specifically, we describe the implementation details of the selected lightweight architecture, the Continual Learning baselines and strategies, and the metrics adopted for evaluation.

\subsection{Implementation Details}
Our experiments focus on CLOD solutions using NanoDet-plus-m~\cite{nanodet}, a lightweight variant of the FCOS object detector. NanoDet, in the selected variant, has a relatively small number of parameters (1.2 million) and a computational footprint of 0.9 Giga-FLOPs, ideal for resource-constrained edge devices~\cite{pastiLD, WANG2025110413}. It is among the smallest models in the literature~\cite{zhang2025few} for CL and few-shot incremental learning applied to object detection. As a reference for its edge suitability, NanoDet achieves inference speeds of 9.8~FPS on a GAP9 MCU and 84~FPS on a Kirin 980 ARM mobile CPU~\cite{bompani}, supporting the feasibility of deploying NanoDet on resource-constrained hardware.

Following common practice in the CLOD literature~\cite{menezes2023continual}, we use a backbone pre-trained on ImageNet, namely ShuffleNet v2~\cite{Ma_2018_ECCV} as implemented in~\cite{nanodet}.
For each task, the models are trained for 50 epochs with a batch size of 32. We use the AdamW optimizer, with a learning rate of 0.001, weight decay of 0.05, a 500-step linear warm-up, and a cosine annealing schedule with $t_{max}$ = 50. 
The validation split of TiROD is used for tuning the training hyperparameter of NanoDet.
Full experimental details appear in Appendix~\ref{sec:appendix}, Table~\ref{tab:hyperparams}.

\subsection{CLOD Methods}
Algorithms in the CLOD literature are usually tailored to specific object detection architectures~\cite{menezes2023continual}. There are no standard frameworks or models, so many state-of-the-art solutions for CLOD cannot be directly applied to any object detection system.
For TiROD, therefore, we selected CLOD techniques that have been successfully applied to FCOS-based architectures such as NanoDet, along with standard continual learning baselines.
Our intent is to establish a clear reference benchmark rather than exhaustively cover all the existing methods, since architectural incompatibilities limit direct comparisons~\cite{menezes2023continual,pastiLD}.

To this end, we evaluate several regularization techniques, including Learning without Forgetting (LwF)\cite{lwf}, Selective and Inter-related Distillation (SID)\cite{peng2021sid}, and IncDet~\cite{liu2020incdet} adapted for FCOS~\cite{peng2023diode}. The distillation coefficient for LwF and SID is set to 1, while for IncDet, the Elastic Weight Consolidation Loss coefficient is 5000, as suggested in~\cite{peng2023diode}.

For replay-based techniques, following common practice~\cite{rolnick2019experience}, we fix the memory buffer to 150 samples, updated so that after each task it contains an equal number of representative images (e.g. 30 per-task on Task 5).
Since TiROD contains video frames, where consecutive images are typically highly correlated rather than i.i.d., we explore three different sampling strategies for updating the replay memory to ensure a diverse and representative selection of images.

The first one is a standard implementation of \textit{Replay}~\cite{replay}, which randomly samples the images from the training task to construct its memory buffer.

Then, motivated by the sequential nature of TiROD's data, we introduce a naive baseline called \textit{Temporal Replay}, which uses linear interpolation to select samples that are as distant in time as possible.
For instance, if the training set consists of 100 sequential frames and 25 frames are needed for the memory buffer, Temporal Replay will select the first frame from each group of~4.

The third method is \textit{K-Means Replay}, introduced in~\cite{yang2021benchmark}. It uses the image pre-logits (i.e., the feature vectors before the classification layer) from a frozen pre-trained to select the buffer images. Following~\cite{yang2021benchmark}, we use an ImageNet-pretrained feature extractor, specifically ResNet-50, to obtain a rich feature space for clustering with $K$ equal to the buffer size. 
The feature extractor is loaded only once per task boundary for buffer selection and is not used during training or inference. 
Clustering is performed on the pre-logits features of images from the current task and the buffer, and the image closest to each centroid is stored.
We also evaluate a lightweight variant of this method using ShuffleNetV2 (which is the backbone of NanoDet) pre-trained on ImageNet as the feature extractor.

Following the CL literature~\cite{menezes2023continual}, we consider the performance of Fine-Tuning as a lower bound for CLOD evaluation, which is trained on task data alone with no technique to avoid catastrophic forgetting.
Similarly, as an upper bound, we consider Cumulative Training, where the model is (re)trained with all the available data from past tasks. Since this method uses all previous data, there is no catastrophic forgetting.

\subsection{Evaluation Metrics}
To evaluate the performances of CLOD algorithms, we consider three metrics.
For all of them, we use the mean Average Precision~(mAP) of the considered CLOD method, weighted at different Intersection over Union~(IoU) from 0.5 to 0.95.

The first metric, \textit{Final mAP}, considers the performance on all tasks after completing the last one.
The other two represent the rates of stability~(\textit{RSD}) and plasticity~(\textit{RPD}) defined as follows:
\begin{align*}
    RSD & = 1 - \frac{1}{N} \sum_{i = 2}^{N} \frac{mAP^{c\_old}_i - mAP^{old}_i}{mAP^{c\_old}_i}\\
    RPD & = 1 - \frac{1}{N} \sum_{i = 2}^{N} \frac{mAP^{c\_new}_i - mAP^{new}_i}{mAP^{c\_new}_i}
\end{align*}
where $mAP^{c\_old}$ (for the Cumulative baseline) and  $mAP^{old}$ (for the evaluated method) refer to the performances on the old tasks $1 \ldots i-1$ after the training on task~$i$.
Similarly, $mAP^{c\_new}_i$ and $mAP^{new}_i$ refer to the performances on the new task $i$ after training on it.
Finally, N is the total number of tasks, in our case 4 for the Intra-Domain and 10 for the Cross-Domain scenarios.

\textit{RSD} quantifies the stability of the model on previously learned tasks across all of them, comparing the Continual Learning's mAP on old data to the mAP achieved with Cumulative Training.
\textit{RPD}, on the other hand, measures the model's plasticity, indicating how well it learns new tasks relative to the Cumulative Training upper bound.
By adopting RSD and RPD, as proposed in the comprehensive CLOD review by Menezes et al.~\cite{menezes2023continual}, we explicitly normalize the performance of the CLOD method against the Cumulative Training upper bound, yielding interpretable percentages that isolate the CLOD methods' ability in retaining old domains (Stability) from their deficit in learning new ones (Plasticity).

\section{TiROD Benchmark Results}\label{sec:experiments}
In this section we present a quantitative analysis of the proposed benchmark. We first discuss the performance of the selected methods on the Cross-Domain and Intra-Domain scenarios, followed by experiments on task ordering and backbone architectures to validate the robustness of our findings.
Finally we evaluate the storage and computational overhead of the CLOD methods studied in the benchmark.

\subsection{Cross-Domain Scenario Results}\label{sec:cross-results}
The Cross-Domain scenario, as defined in Sec.~\ref{sec:tirod_dataset_benchmarks}, represents the most challenging of the two benchmarks considered, as it combines both domain and class-incremental learning across tasks.

As shown in Table~\ref{tab:nanodet}, regularization methods generally underperformed compared to replay-based techniques, suggesting that they are less effective when tasks involve significant domain shifts. Among them, SID was the best one, achieving a Final mAP of $20.8 \pm 2.42\%$. IncDet and LwF showed moderate stability but higher plasticity, indicating they effectively learn new tasks but struggle with catastrophic forgetting. Furthermore, these regularization techniques exhibited relatively high variance across random seeds. Latent Distillation performed similarly to SID, but offered better efficiency by freezing and sharing the backbone between teacher and student~\cite{pastiLD}.

\begin{table}
    \caption{TiROD Cross-Domain scenario results. Performance is reported as the mean $\pm$ standard deviation across three random seeds.}
    \centering
    \label{tab:nanodet}
    \footnotesize
    \begin{tabular}{@{}lllll@{}}
    \toprule
    \textbf{Method} &\textbf{Final mAP [\%]} & \textbf{RSD} $\mathbf{\uparrow}$ & \textbf{RPD} $\mathbf{\uparrow}$ \\
    \midrule
    \midrule
    Fine-Tuning & 13.4 $\pm$ 1.34 & 0.16 $\pm$ 0.01 & 0.99 $\pm$ 0.01 \\
    \midrule
    LwF & 16.2 $\pm$ 2.50 & 0.22 $\pm$ 0.03 & 0.98 $\pm$ 0.01\\
    IncDet & 12.9 $\pm$ 0.60 & 0.32 $\pm$ 0.09 & 0.90 $\pm$ 0.03 \\
    SID & 20.8 $\pm$ 2.42 & 0.39 $\pm$ 0.01 & 0.87 $\pm$ 0.02 \\
    \midrule
    Replay & 38.4 $\pm$ 0.74 & 0.69 $\pm$ 0.01 & 0.89 $\pm$ 0.09 \\
    Temporal Replay & 27.3 $\pm$ 1.14 & 0.47 $\pm$ 0.02 & 0.96 $\pm$ 0.01 \\
    K-Means Replay (ResNet) & \textbf{42.7 $\pm$ 0.20} & 0.75 $\pm$ 0.01 & 0.97 $\pm$ 0.01 \\
    K-Means Replay (ShuffleNet) & 39.7 $\pm$ 1.06 & 0.73 $\pm$ 0.02 & 0.98 $\pm$ 0.01\\
    \midrule
    Latent Distillation & 19.8 $\pm$ 2.90 & 0.30 $\pm$ 0.05 & 0.81 $\pm$ 0.03\\
    Latent Replay & 36.5 $\pm$ 0.7 & 0.63 $\pm$ 0.02 & 0.93 $\pm$ 0.02 \\
    Latent K-Means Replay (ResNet) & 39.1 $\pm$ 1.40 & 0.70 $\pm$ 0.02 & 0.94 $\pm$ 0.03 \\
    Latent K-Means Replay (ShuffleNet) & 38.5 $\pm$ 1.20 & 0.72 $\pm$ 0.01 & 0.94 $\pm$ 0.01 \\
    \midrule
    \midrule
    Cumulative Training & 63.2 $\pm$ 0.34 & - & - \\
    \bottomrule
    \end{tabular}
\end{table}


Replay-based techniques consistently outperformed all regularization methods. Temporal Replay was the worst of them because, when the robot was still and the scene unchanged, it selected too many unrepresentative frames, which highlights the need to develop more effective sampling techniques. Standard and Latent Replay, both using random sampling, achieved a Final mAP of $38.4 \pm 0.74\%$ and $36.5 \pm 0.70\%$, respectively. While Latent Replay is more computationally efficient with the frozen backbone, it showed slightly worse plasticity, since updating just the upper layers limited the adaptation to new tasks. 

Finally, K-Means Replay achieved the best overall performance across all CLOD techniques. The ResNet variant yielded a Final mAP of $42.7 \pm 0.20\%$, showing extreme stability across runs and confirming the advantage of selecting more representative samples. The ShuffleNet variant achieved $39.7 \pm 1.06\%$, representing a minor trade-off in accuracy for a good reduction in deployment memory footprint.

\begin{figure}
    \centering
    \includegraphics[width=\linewidth]{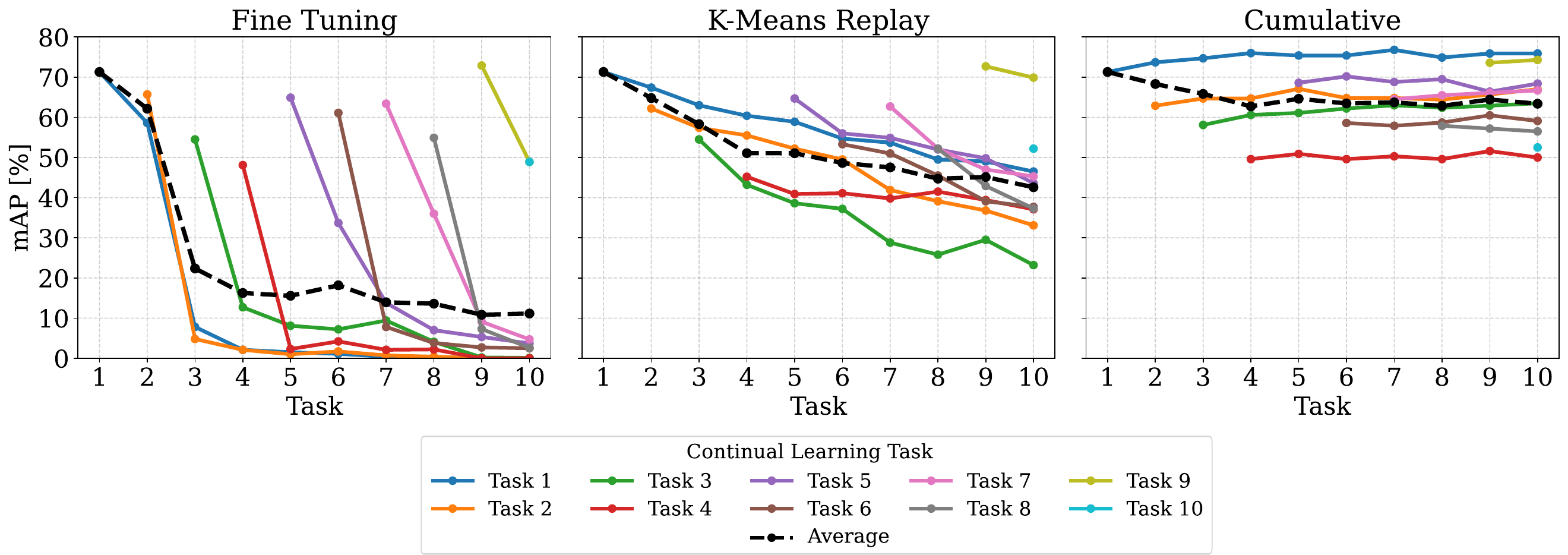}
    \caption{\emph{TiROD} per-task Mean Average Precision (mAP) for the Cross-Domain scenario. We show the performances of Fine-Tuning and Cumulative as the lower and upper bounds, and K-Means Replay (ResNet) as the best performing CL approach in Table~\ref{tab:nanodet}. At each training step $i$, the curves show the model's mAP on all tasks $1 \ldots i$ after completing training on task $i$. Different colors correspond to different test tasks. The dashed black line shows the average mAP across all evaluated tasks up to each training step.}\label{fig:results}
\end{figure}

Although the replay-based methods have partial access to previous data, all the CLOD techniques perform training only on the current task's data, leading to a Final mAP lower than the case where training is done Cumulatively on all the available data. This trend is also evident in Fig.~\ref{fig:results}, where the per-task mAP of the best performing CLOD method (K-Means Replay (ResNet)) is compared against the Fine-Tuning and Cumulative Training's lower- and upper-bound. Replay partially mitigates catastrophic forgetting, but a significant gap remains, which motivates further research in CLOD architectures.

To further evaluate robustness, we repeated the Cross-Domain experiments under two alternative task orders (reverse domains, reverse illumination and randomized). Our results, reported in Table~\ref{tab:task_order_mean}, show a consistent trend, with K-Means Replay achieving again the best performance among all the CLOD strategies.

We also analyze the influence of the buffer size on the best-performing replay technique. As reported in Table~\ref{tab:buffer}, even under tighter constraints, K-Means Replay outperforms all regularization methods. However, as expected, reducing the buffer size leads to a gradual decrease in performance, so further improvements would be useful to operate satisfactorily under extreme memory and computation constraints.

\begin{table}
    \caption{Replay buffer size impact on K-Means Replay (ResNet) in the Cross-Domain scenario (see also Table~\ref{tab:nanodet}).}
    \centering
    \label{tab:buffer}
    \footnotesize
    \begin{tabular}{@{}lllll@{}}
    \toprule
    \textbf{Method} &\textbf{Final mAP [\%]} & \textbf{RSD} $\mathbf{\uparrow}$ & \textbf{RPD} $\mathbf{\uparrow}$ \\
    \midrule
    \midrule
    K-Means Replay (ResNet) (buf=150) & \textbf{42.5} & 0.77 & 0.95 \\
    K-Means Replay (ResNet) (buf=100) & 36.4 & 0.69 & 0.97 \\
    K-Means Replay (ResNet) (buf=50) & 27.4 & 0.60 & 0.97 \\
    K-Means Replay (ResNet) (buf=25) & 21.9 & 0.49 & 0.98 \\
    \bottomrule
    \end{tabular}
\end{table}

\subsection{Intra-Domain Scenario Results}\label{sec:intra-results}
\begin{table}
\caption{TiROD Intra-Domain scenario results. Performance is reported as the mean $\pm$ standard deviation across three random seeds.}
\centering
\label{tab:nanodet-intra}
\footnotesize
\begin{tabular}{@{}lllll@{}}
\toprule
\textbf{Method} &\textbf{Final mAP [\%]} & \textbf{RSD} $\mathbf{\uparrow}$ & \textbf{RPD} $\mathbf{\uparrow}$ \\
\midrule
\midrule
Fine-Tuning & 46.2 $\pm$ 0.56 & 0.79 $\pm$ 0.2  & 0.94 $\pm$ 0.01 \\
\midrule
LwF & 46.5 $\pm$ 1.81 & 0.77 $\pm$ 0.01 & 0.92 $\pm$ 0.01\\
IncDet & 42.9 $\pm$ 2.13 & 0.81 $\pm$ 0.04 & 0.71 $\pm$ 0.16 \\
SID & 47.8 $\pm$ 0.90 & 0.75 $\pm$ 0.04 & 0.88  $\pm$ 0.02 \\
\midrule
Replay & 66.2 $\pm$ 0.25 & 0.94 $\pm$ 0.01 & 0.94 $\pm$ 0.02\\
Temporal Replay & 60.5 $\pm$ 0.07 & 0.88 $\pm$ 0.01 & 0.92 $\pm$ 0.01\\
K-Means Replay (ResNet) & 66.5 $\pm$ 0.66 & 0.95 $\pm$ 0.01 & 0.95 $\pm$ 0.01 \\
K-Means Replay (ShuffleNet) & \textbf{67.4 $\pm$ 0.81} & 0.93 $\pm$ 0.01 & 0.96 $\pm$ 0.01 \\
\midrule
Latent Distillation & 50.1 $\pm$ 3.81 & 0.61 $\pm$ 0.01 & 0.81 $\pm$ 0.01 \\
Latent Replay & 64.4 $\pm$ 1.49 & 0.92 $\pm$ 0.01 & 0.91 $\pm$ 0.01\\
Latent K-Means Replay (ResNet) & 65.8 $\pm$ 0.60 & 0.92 $\pm$ 0.01 & 0.91 $\pm$ 0.01 \\
Latent K-Means Replay (ShuffleNet) & 65.9 $\pm$ 1.57 & 0.91 $\pm$ 0.01 & 0.91 $\pm$ 0.01 \\
\midrule
\midrule
Cumulative Training & 70.5 $\pm$ 0.41  & - & - \\
\bottomrule
\end{tabular}
\end{table}

The Intra-Domain scenario, introduced in Sec.~\ref{sec:tirod_dataset_benchmarks}, focuses on appearance shifts within a single environment, including changes in illumination and weather conditions, while the class remains the same.
As reported in Table~\ref{tab:nanodet-intra}, we observe a trend similar to the Cross-Domain scenario; indeed, Replay-based methods consistently outperform regularization-based approaches.

Among the regularization methods, SID achieved the highest Final mAP ($47.8 \pm 0.90\%$), closely followed by LwF. These approaches generally preserved better RSD and RPD scores compared to the Cross-Domain setting, suggesting that intra-domain appearance shifts are easier to manage than full domain shifts, even without a replay memory. However, the improvement with respect to standard Fine-Tuning ($46.2 \pm 0.56\%$) is only marginal.

Replay-based methods led again to the best results overall, although in this case K-Means Replay only slightly outperformed the other methods, indicating that under appearance shifts the memory selection mechanism is less effective. Note also that ShuffleNet variant of K-Means Replay performs comparably to the ResNet version, suggesting that a lighter feature extractor suffices in the Intra-Domain scenario.

As expected, Cumulative Training ($70.5 \pm 0.41\%$) still outperformed all CL strategies. However, the performance gap between the cumulative upper bound and the best CLOD strategy ($\sim 3.1\%$) is much less pronounced here than in the Cross-Domain scenario, demonstrating that current replay techniques are already highly capable of addressing intra-domain weather and lighting shifts.

\subsection{Alternative Task Orders in Cross-Domain Benchmark}
To further evaluate the reliability of our benchmark with respect to task ordering, we repeat the Cross-Domain experiments under four alternative orderings.

The first is the original task order of the Cross-Domain benchmark reported in Sec.~\ref{sec:tirod_dataset_benchmarks} (Domain 1 High to Domain 5 Low).
The second one is a reversed order, starting from the last domain (Domain 5 Low) and proceeding backwards to Domain 1 High. 
The third is a randomized order \textit{(d3\_h, d3\_l, d5\_h, d5\_l, d1\_h, d1\_l, d2\_h, d2\_l, d4\_h, d4\_l)}.
The fourth swaps the illumination order within each domain, proceeding from Low to High illumination \textit{(d1\_l, d1\_h, d2\_l, d2\_h, \ldots, d5\_l, d5\_h)}.

We report in Table~\ref{tab:task_order_mean} the aggregate results over the four task orderings, with the mean and standard deviation for each metric.
Overall, the relative ranking of methods remains consistent with the main results in Table~\ref{tab:nanodet}. Replay-based methods consistently outperform regularization-based ones, with K-Means Replay providing the best overall performance across all task orders.
This confirms that the benchmark findings are robust to task order variations, mitigating the concern that results may be specific to a specific CL sequence.

\begin{table}
\centering
\caption{TiROD Cross-Domain scenario results with mean $\pm$ standard deviation across four task orders (original, reversed, randomized, and low-to-high illumination).}
\label{tab:task_order_mean}
\footnotesize
\begin{tabular}{@{}lccc@{}}
\toprule
\textbf{Method} & \textbf{Final mAP [\%]} & \textbf{RSD} $\uparrow$ & \textbf{RPD} $\uparrow$ \\
\midrule
Fine-Tuning         & $12.9 \pm 1.7$ & $0.23 \pm 0.08$ & $0.98 \pm 0.01$ \\
LwF                 & $14.9 \pm 1.8$ & $0.29 \pm 0.07$ & $0.98 \pm 0.01$ \\
IncDet              & $12.7 \pm 2.2$ & $0.27 \pm 0.08$ & $0.88 \pm 0.09$ \\
SID                 & $19.5 \pm 2.8$ & $0.45 \pm 0.05$ & $0.86 \pm 0.02$ \\
\midrule
Replay              & $38.1 \pm 1.7$ & $0.71 \pm 0.05$ & $0.94 \pm 0.01$ \\
Temporal Replay     & $29.3 \pm 2.4$ & $0.53 \pm 0.04$ & $0.95 \pm 0.02$ \\
K-Means Replay (ResNet) & $\mathbf{40.4 \pm 2.3}$ & $0.74 \pm 0.03$ & $0.95 \pm 0.02$ \\
K-Means Replay (ShuffleNet) & $39.3 \pm 0.5$ & $0.72 \pm 0.02$ & $0.96 \pm 0.04$ \\
\midrule
Latent Distillation & $18.9 \pm 4.2$ & $0.34 \pm 0.03$ & $0.80 \pm 0.03$ \\
Latent Replay       & $37.5 \pm 0.7$ & $0.71 \pm 0.04$ & $0.91 \pm 0.02$ \\
Latent K-Means Replay (ResNet)     & $39.5 \pm 1.8$ & $0.71 \pm 0.03$ & $0.92 \pm 0.02$ \\
Latent K-Means Replay (ShuffleNet)     & $37.8 \pm 2.0$ & $0.71 \pm 0.02$ & $0.91 \pm 0.03$ \\
\midrule
Cumulative Training & $63.2 \pm 0.3$ & - & - \\
\bottomrule
\end{tabular}
\end{table}

\subsection{Per-Class Performance Dynamics}
\begin{figure}
    \centering
\includegraphics[width=\textwidth]{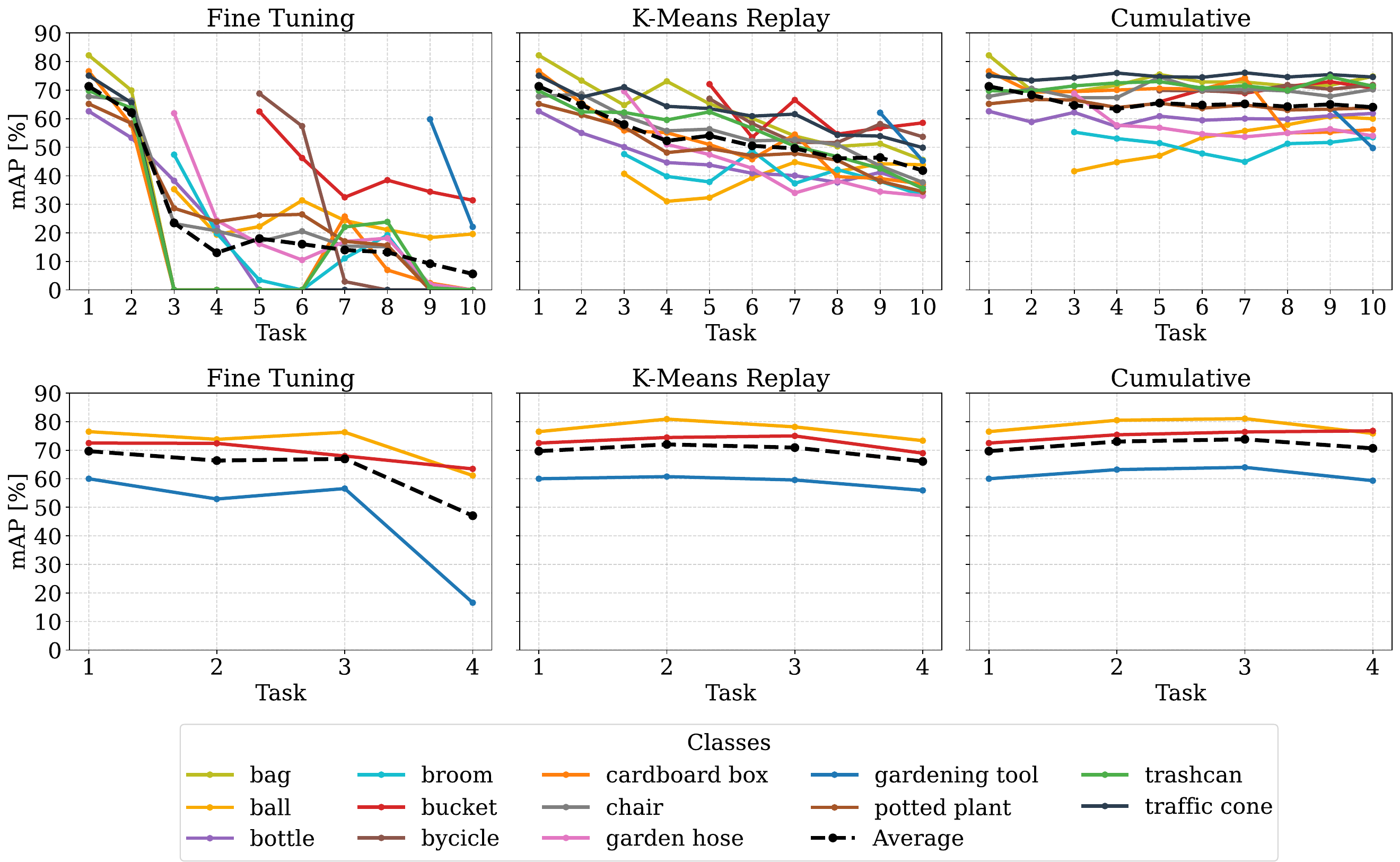}
    \caption{\emph{TiROD} per class category mAP for the learning tasks of the Cross Domain scenario (top) and Intra-Domain scenario (bottom). We show the performances of Fine-Tuning and Cumulative as the lower and upper bounds, and K-Means Replay (ResNet) as the best performing CL approach in Table~\ref{tab:nanodet}. The dashed black line shows the average mAP across all classes at each task.\label{fig:classResults}}
\end{figure}

We also report per-class mAP [ IoU $=0.5:0.95$ ] for the Cross-Domain scenario in Fig.~\ref{fig:classResults}~(top), comparing the best performing CL method (K-Means Replay (ResNet)) against the Fine-Tuning and Cumulative bounds.
These plots complement the global per-task curves in Fig.~\ref{fig:results} and help distinguish the CIL versus DIL challenges of TiROD. While the Cumulative baseline remains largely stable across tasks, Fine-Tuning typically suffers strong catastrophic forgetting on classes encountered early, while Replay substantially reduces this effect.

Per-class performances reveal two distinct phenomena. First, although for some categories the mAP can increase when a class is seen under different conditions (e.g. \emph{ball}), the overall trend is still dominated by the forgetting of previous classes. Second, classes that reappear in later tasks (e.g. \emph{cardboard box} in Tasks 1–2 and then again in Task 7) often show a partial mAP recovery when re-observed, both with Fine-Tuning and K-Means Replay. However, their impact on the overall mAP is marginal compared to the catastrophic forgetting of classes that do not reappear.

Similarly, we analyze the per-class trajectories for the Intra-Domain scenario in Fig.~\ref{fig:classResults}~(bottom). Interestingly, in some cases, observing the same classes under intra-domain variations yields a small boost in precision even with simple Fine-Tuning (e.g. \emph{bucket} from Task~1 to Task~2, and \emph{ball} from Task~2 to Task~3). However, in general, Replay-based methods such as K-Means Replay provide greater robustness against catastrophic forgetting, particularly when larger appearance shifts occur (e.g. Task~3 to Task~4).

\subsection{Different Backbones Results}
To further validate our results, we extend the Cross-Domain Benchmark analysis by comparing the best replay and the best regularization methods from Table~\ref{tab:nanodet}, namely SID and K-Means Replay, with three bigger NanoDet backbones: ShuffleNetv2-1.5x, MobileNetV2, and EfficientNet-Lite~\cite{nanodet}. These backbones increase the total number of the detection model parameters to 2.4M, 2.7M, and 4.7M, respectively.
As per the main experiments, the backbones were pre-trained on ImageNet, and we applied the same training schedule with 50 epochs per CL task.

As reported in Table~\ref{tab:backbone_cmp}, the relative performance trends remain consistent to the results obtained with the baseline version of ShuffleNetv2. Replay-based techniques substantially mitigate catastrophic forgetting compared to Fine-Tuning, while regularization methods only achieve marginal improvements. This consistency reinforces our finding that Replay is a more robust strategy for handling the severe domain shifts present in TiROD's Cross-Domain scenario.
We also note that larger backbones mAPs do not improve much with respect to the baseline, highlighting the fact that TiROD remains a challenging benchmark even for bigger models.

\begin{table}
\caption{TiROD Cross-Domain results across backbones.}
\centering
\label{tab:backbone_cmp}
\footnotesize
\setlength{\tabcolsep}{2pt}
\begin{tabular}{@{}llccc@{}}
\toprule
\textbf{Backbone} & \textbf{Method} & \textbf{Final mAP [\%]} & \textbf{RSD} $\uparrow$ & \textbf{RPD} $\uparrow$ \\
\midrule
\multirow{4}{*}{\shortstack[l]{ShuffleNetV2\\(1.2M params)}} 
& Fine-Tuning     & 10.7 & 0.17 & 0.97 \\
& SID             & 16.4 & 0.41 & 0.84 \\
& K-Means Replay  & 42.5 & 0.77 & 0.95 \\
& Cumulative      & 63.0 & -    & -    \\
\midrule
\multirow{4}{*}{\shortstack[l]{ShuffleNetV2-1.5x\\(2.4M params)}} 
& Fine-Tuning     & 13.1 & 0.25 & 0.99 \\
& SID             & 19.1 & 0.49 & 0.91 \\
& K-Means Replay  & 42.8 & 0.78 & 0.96 \\
& Cumulative      & 63.6 & -    & -    \\
\midrule
\multirow{4}{*}{\shortstack[l]{MobileNetV2\\(2.7M params)}} 
& Fine-Tuning     & 10.1 & 0.18 & 0.99 \\
& SID             & 11.5 & 0.31 & 0.97 \\
& K-Means Replay  & 29.1 & 0.68 & 0.91 \\
& Cumulative      & 61.1 & -    & -    \\
\midrule
\multirow{4}{*}{\shortstack[l]{EfficientNet-Lite\\(4.7M params)}} 
& Fine-Tuning     & 14.1 & 0.27 & 0.99 \\
& SID             & 16.7 & 0.48 & 0.91 \\
& K-Means Replay  & 39.8 & 0.72 & 0.92 \\
& Cumulative      & 63.7 & -    & -    \\
\bottomrule
\end{tabular}
\end{table}

\subsection{Computational and Memory Overhead Analysis}
\label{sec:overhead}
While the previous sections evaluated the Continual Learning capabilities of CLOD methods, directly deploying these solutions on tiny mobile robots introduces strict constraints on memory footprint, compute, and energy consumption that depend on specific hardware platforms.
To this end, we assess the computational requirements of the benchmarked CLOD methods with a theoretical overhead analysis reported in Table~\ref{tab:overhead}. 
For each method, we derive the additional disk storage, training RAM, and compute required relative to a standard Fine-Tuning baseline. 
Because these quantities depend only on the model architecture and CLOD method, they provide general guidelines on the hardware-agnostic bounds broadly applicable to resource-constrained deployment scenarios.

All methods share the same base detector, NanoDet-Plus-m (${\sim}$1.2\,M parameters, 0.9\,GFLOPs), whose checkpoint occupies 4.7\,MB at FP32 precision. 
Storage overhead for replay-based methods assumes a buffer of 150 compressed PNG images at the average TiROD file size of ${\sim}$0.19\,MB each. 
Training RAM overhead captures the additional (or saved) memory due to extra model copies, optimizer states, and gradient tensors. Peak RAM refers to the peak memory required by the K-Means Replay methods in the buffer selection stage. Note that activation memory scales with batch size and input resolution and is identical across all methods and is therefore omitted from the comparison.
Finally, compute scaling refers to the ratio of training computations per epoch with respect to standard Finetuning

\begin{table}[h!]
\centering
\caption{Theoretical compute and memory overhead of each CLOD method relative to a standard Fine-Tuning baseline.}
\label{tab:overhead}
\setlength{\tabcolsep}{4pt}
\newcommand{\tb}[1]{#1}
\begin{tabular}{l c c c c}
\toprule
\textbf{Method}
  & \makecell{\textbf{Storage} \\ \textbf{(MB)}}
  & \makecell{\textbf{Train RAM} \\ \textbf{(MB)}}
  & \makecell{\textbf{Peak RAM} \\ \textbf{(MB)}}
  & \textbf{Compute} \\
\midrule
Fine-Tuning (Baseline) & --  & --   & --     & 1.00$\times$ \\
\midrule
LwF & +4.7  & +4.7  & --     & 1.33$\times$ \\
SID & +4.7  & +4.7  & --     & 1.33$\times$ \\
IncDet & +9.4  & +9.4  & --     & 1.00$\times$ \\
\midrule
Replay & +28.8 & 0.0   & --     & 2.00$\times$ \\
K-Means Replay (ResNet) & +131.0& 0.0   & +102.2 & 2.00$\times$ \\
\tb{K-Means Replay (ShuffleNet)} & \tb{+33.8} & \tb{0.0}   & \tb{+3.2}   & \tb{2.00$\times$} \\
\midrule
Latent Distillation & +1.6  & $-$4.6& --     & 0.77$\times$ \\
Latent Replay & +28.8 & $-$6.2& --     & 1.25$\times$ \\
Latent K-Means Replay (ResNet) & +131.0& $-$6.2& +102.2 & 1.25$\times$ \\
\tb{Latent K-Means Replay (ShuffleNet)} & \tb{+33.8} & \tb{$-$6.2}& \tb{+3.2}   & \tb{1.25$\times$} \\
\bottomrule
\multicolumn{5}{p{\columnwidth}}{\footnotesize
}
\end{tabular}
\end{table}

Distillation methods like LwF and SID require a full frozen model copy to act as the teacher, increasing both storage and training RAM, while adding a 1.33$\times$ compute overhead for the additional teacher forward pass.
IncDet maintains baseline compute but incurs a +9.4,MB overhead in both disk storage and training RAM. This accounts for the Fisher Information matrix used to estimate parameter importance, plus a frozen copy of the previous model's weights required for the elastic regularization penalty.

Replay methods, while performing well in the benchmark, double the compute as for each new training sample, the model also trains on a random sample from the memory buffer.
K-Means Replay introduces a transient peak RAM of +102.2\,MB during the buffer update phase. Replacing ResNet-50 with ShuffleNetV2 mitigates this bottleneck, reducing the peak RAM to just +3.2\,MB with a small drop in performances in the Cross-Domain scenario.

Latent methods offer an interesting trade-off for resource-constrained platforms. By freezing the backbone, they eliminate its gradient and optimizer state buffers, require less training memory than standard Fine-Tuning. Latent Distillation achieves the lowest compute scaling (0.77$\times$) as it shares the frozen backbone between Teacher and Student.
Latent versions of Replay balance training RAM memory savings ($-$6.2\,MB) with a moderate compute overhead (1.25$\times$). Ultimately, while K-Means Replay (ResNet) achieves the highest absolute accuracy in our benchmarks, Latent K-Means Replay (ShuffleNet) offers a good compromise, sacrificing some accuracy, but requiring substantially less resources.

Note that on an RTX~2080~Ti GPU, a single TiROD Fine-Tuning task completes in approximately 5 minutes (50 epochs, batch size~32). The compute scaling factors in Table~\ref{tab:overhead} allow estimating relative training durations for each CLOD strategy: for instance, Replay at 2.00$\times$ requires ${\sim}$10 minutes per task, while Latent Distillation at 0.77$\times$ requires less than 4 minutes. While training these models directly on current microcontrollers remains a challenge, these hardware-agnostic bounds show that Latent methods offer the most viable path toward on-device continual adaptation.

\section{Conclusions and Limitations}\label{sec:conclusions}
In this paper, we introduce the TiROD dataset along with a comprehensive benchmark for evaluating CLOD systems on resource-constrained mobile robots. The dataset was collected across multiple domains, indoors and outdoors, providing a testbed for assessing real-world challenges such as domain shifts, motion blur, and noisy sensor data, conditions commonly encountered in Tiny Robotics applications. 
We evaluated various CLOD techniques in two different scenarios using NanoDet, a lightweight object detector. The results showed that Replay-based approaches, especially when using representative sampling strategies, consistently outperformed regularization methods. This suggests the former can better adapt to new environments without catastrophically forgetting the old ones, at least when past data is available.

However, the performance gap with respect to Cumulative Training shows that the current methods for Continual Learning are still far from optimal in real-world and resource-constrained scenarios with significant domain shifts. New approaches are therefore needed to close this gap, especially for tiny autonomous robots with limited memory resources.

While TiROD introduces a novel controlled research benchmark for CLOD in the tiny robotics domain, its current scale remains limited. To build the robustness required for real-world applications across diverse sites, weather conditions, and sensors, we welcome community contributions of compatible datasets (requiring COCO-style annotations and hardware specifications such as camera type and robot platform), aiming to collaboratively expand the benchmark's scale and diversity.

Finally, while this benchmark focuses strictly on vision-based Continual Learning, future work could also explore multi-modal data selection strategies by incorporating additional sensors, such as IMUs or wheel odometry, to enable smarter replay buffer management.
\newpage

\begin{appendices}

\section{Additional Experimental Details}
\label{sec:appendix}
We summarise the key hyperparameters used for the experiments in Table~\ref{tab:hyperparams}.
In particular, the training hyperparameters used for NanoDet were tuned using the validation split of TiROD.
The same training settings were applied across all CL tasks and methods. For techniques requiring additional loss terms (e.g., SID, IncDet), we adopted values consistent with prior work. Data augmentation follows the configuration used in~\cite{nanodet}. All experiments were conducted on an NVIDIA GeForce RTX 2080Ti GPU with 12 GB of RAM.

\begin{table}[h]
\caption{Hyperparameter settings.}
\centering
\label{tab:hyperparams}
\footnotesize
\begin{tabular}{@{}ll@{}}
\toprule
\textbf{Parameter} & \textbf{Value} \\
\midrule
Batch size & 32 \\
Epochs per CL task & 50 \\
Optimizer & AdamW \\
Learning rate & 0.001 \\
Weight decay & 0.05 \\
Cosine Annealing schedule &  $t_{max}=50$ \\
Warm-up steps & 500 linear \\
Replay buffer size & 150 images \\
Bbox loss weight & $\alpha=2$ \\
Distribution Focal loss weight & $\alpha=0.25$ \\
Quality Focal loss weights & $\alpha=1.0$, $\beta=2.0$ \\
LwF distillation weight & $\alpha=1$ \\
SID distillation weights & $\alpha=1$, $\beta=1$ \\
IncDet EWC loss coeff. & $\lambda=5000$ \\
\bottomrule
\end{tabular}
\end{table}

\end{appendices}

\newpage
\bibliography{sn-bibliography}

\end{document}